\documentclass[10pt,twocolumn,letterpaper]{article}

\usepackage[pagenumbers]{cvpr} 
\usepackage{graphicx}
\usepackage{amsmath}
\usepackage{amssymb}
\usepackage{booktabs}
\usepackage{bm}
\usepackage[ruled,linesnumbered]{algorithm2e}
\usepackage{pslatex}
\usepackage[accsupp]{axessibility}  

\usepackage[pagebackref,breaklinks,colorlinks]{hyperref}
\usepackage{marvosym}

\usepackage[capitalize]{cleveref}
\crefname{section}{Sec.}{Secs.}
\Crefname{section}{Section}{Sections}
\Crefname{table}{Table}{Tables}
\crefname{table}{Tab.}{Tabs.}

\usepackage{booktabs}                                   
\usepackage{multirow}                                   
\usepackage{makecell}                                   
\usepackage{tablefootnote}                              
\usepackage[symbol]{footmisc}                           
\usepackage{amsmath,amssymb}                            
\usepackage{xcolor}                                     
\usepackage{enumitem}                                   
\usepackage{subcaption}                                 

\usepackage{hyperref}   

\newcommand{\para}[1]{\paragraph{#1}}

\def\cpt{CP$^3$}

\def\cvprPaperID{9129} 
\def\confName{CVPR}
\def\confYear{2023}

\begin{document}

\title
{CP$^3$: Channel Pruning Plug-in for Point-based Networks}
\author{Yaomin Huang\textsuperscript{1,2,*}
\and
Ning Liu\textsuperscript{2,*}
\and
Zhengping Che\textsuperscript{2}
\and
Zhiyuan Xu\textsuperscript{2}
\and
Chaomin Shen\textsuperscript{1}
\and
Yaxin Peng\textsuperscript{3}
\and
Guixu Zhang\textsuperscript{1,\Letter}
\and
Xinmei Liu\textsuperscript{1}
\and
Feifei Feng\textsuperscript{2}
\and
Jian Tang\textsuperscript{2,\Letter}
\and
\textsuperscript{1}School of Computer Science, East China Normal University \\
\textsuperscript{2}Midea Group \\
\textsuperscript{3}Department of Mathematics, School of Science, Shanghai University\\
{\tt\small \textsuperscript{1}\{51205901049,51205901078\}@stu.ecnu.edu.cn \quad \textsuperscript{1}\{cmshen,gxzhang\}@cs.ecnu.edu.cn} \\
{\tt\small \textsuperscript{2}\{liuning22,chezp,xuzy70,feifei.feng,tangjian22\}@midea.com \quad \textsuperscript{3}yaxin.peng@shu.edu.cn}
}

\maketitle
\footnote[0]{* Equal contributions.}
\footnote[0]{\textsuperscript{\Letter} Corresponding authors.}
\footnote[0]{This work is done during Yaomin Huang and Xinmei Liu's internship at Midea Group.}

\begin{abstract}
Channel pruning can effectively reduce both computational cost and memory footprint of the original network while keeping a comparable accuracy performance.
Though great success has been achieved in channel pruning for 2D image-based convolutional networks (CNNs), existing works seldom extend the channel pruning methods to 3D point-based neural networks~(PNNs).
Directly implementing the 2D CNN channel pruning methods to PNNs undermine the performance of PNNs because of the different representations of 2D images and 3D point clouds as well as the network architecture disparity.
In this paper, we proposed {\cpt}, which is a \textbf{C}hannel \textbf{P}runing \textbf{P}lug-in for \textbf{P}oint-based network.
{\cpt} is elaborately designed to leverage the characteristics of point clouds and PNNs in order to enable 2D channel pruning methods for PNNs.
Specifically, it presents a coordinate-enhanced channel importance metric to reflect the correlation between dimensional information and individual channel features,
and it recycles the discarded points in PNN's sampling process and reconsiders their potentially-exclusive information to enhance the robustness of channel pruning.
Experiments on various PNN architectures show that CP$^3$ constantly improves state-of-the-art 2D CNN pruning approaches on different point cloud tasks.
For instance, our compressed \mbox{PointNeXt-S} on ScanObjectNN  achieves an accuracy of 88.52\% with a pruning rate of 57.8\%, outperforming the baseline pruning methods with an accuracy gain of 1.94\%.

\end{abstract}

\section{Introduction}
\label{sec:intro}
Convolutional Neural Networks (CNNs) often encounter the problems of overloaded computation and overweighted storage.
The cumbersome instantiation of a CNN model leads to inefficient, uneconomic, or even impossible deployment in practice.
Therefore, light-weight models that provide comparable results with much fewer computational costs are in great demand for nearly all applications.
Channel pruning is a promising solution to delivering efficient networks.
In recent years, 2D CNN channel pruning, e.g., pruning classical VGGNets~\cite{simonyan2014very}, ResNets~\cite{he2016deep}, MobileNets~\cite{howard2017mobilenets}, and many other neural networks for processing 2D images~\cite{liu2017learning,ding2019centripetal,lin2020hrank,meng2020pruning,ding2021resrep,sui2021chip,guan2022dais}, has been successfully conducted.
Most channel pruning approaches focus on identifying redundant convolution filters (i.e., channels) by evaluating their importance.
The cornerstone of 2D channel pruning methods is the diversified yet effective channel evaluation metrics.
For instance, HRank~\cite{lin2020hrank} uses the rank of the feature map as the pruning metric and removes the low-rank filters that are considered to contain less information.
CHIP~\cite{sui2021chip} leverages channel independence to represent the importance of each feature mapping and eliminates less important channels.

With the widespread application of depth-sensing technology, 3D vision tasks~\cite{gong2022posetriplet,10.1007/978-3-031-19824-3_41,DBLP:conf/cvpr/FazlaliXRL22,DBLP:conf/cvpr/WangWCXQQFT22} are a rapidly growing field starving for powerful methods.
Apart from straightforwardly applying 2D CNNs, models built with Point-based Neural Networks~(PNNs), which directly process point clouds from the beginning without unnecessary rendering,
show their merits and are widely deployed on edge devices for various applications such as robots~\cite{DBLP:conf/iros/YangPCF20, li2022efficientgrasp} and self-driving~\cite{DBLP:conf/cvpr/ZhengTJF21,DBLP:conf/aaai/ChenCZT22}.
Compressing PNNs is crucial due to the limited resources of edge devices and multiple models for different tasks are likely to run simultaneously~\cite{Pham_2019_CVPR, DUBEY2022100275}.
Given the huge success of 2D channel pruning and the great demand for efficient 3D PNNs, we intuitively raise one question: \emph{shall we directly implement the existing pruning methods to PNNs following the proposed channel importance metrics in 2D CNNs pruning?}

With this question in mind, we investigate the fundamental factors that potentially impair 2D pruning effectiveness on PNNs.
Previous works~\cite{koppula2011semantic,xu2021image2point} have shown that point clouds record visual and semantic information in a significantly different way from 2D images.
Specifically, a point cloud consists of a set of unordered points on objects' and environments' surfaces, and each point encodes its features,
such as intensity along with the spatial coordinates $(x, y, z)$.
In contrast, 2D images organize visual features in a dense and regular pixel array.
Such data representation differences between 3D point clouds and 2D images lead to a) different ways of exploiting information from data and b) contrasting network architectures of PNNs and 2D CNNs.
It is credible that only the pruning methods considering the two aspects (definitely not existing 2D CNN pruners) may obtain superior performance on PNNs.

From the perspective of data representations,
3D point clouds provide more 3D feature representations than 2D images, but the representations are more sensitive to network channels.
To be more specific, for 2D images, all three RGB channels represent basic information in an isotropic and homogeneous way so that the latent representations extracted by CNNs applied to the images.
On the other hand, point clouds explicitly encode the spatial information in three coordinate channels, which are indispensable for extracting visual and semantic information from other channels.
Moreover, PNNs employ the coordinate information in multiple layers as concatenated inputs for deeper feature extraction.
Nevertheless, existing CNN pruning methods are designed only suitable for the plain arrangements of 2D data but fail to consider how the informative 3D information should be extracted from point clouds.

Moreover, the network architectures of PNNs are designed substantially different from 2D CNNs.
While using smaller kernels~\cite{simonyan2014very} is shown to benefit 2D CNNs~\cite{simonyan2014very}, it does not apply to networks for 3D point clouds.
On the contrary, PNNs leverage neighborhoods at multiple scales to obtain both robust and detailed features.
The reason is that small neighborhoods (analogous to small kernels in 2D CNNs) in point clouds consist of few points for PNNs to capture robust features.
Due to the necessary sampling steps, the knowledge insufficiency issue becomes more severe for deeper PNN layers.
In addition, PNNs use the random input dropout procedure during training to adaptively weight patterns detected at different scales and combine multi-scale features.
This procedure randomly discards a large proportion of points and loses much exclusive information of the original data.
Thus, the architecture disparity between 2D CNNs and PNNs affects the performance of directly applying existing pruning methods to PNNs.

In this paper, by explicitly dealing with the two characteristics of 3D task, namely the data representation and the PNN architecture design,
we propose a \textbf{C}hannel \textbf{P}runing \textbf{P}lug-in for \textbf{P}oint-based network named {\cpt},
which can be applied to most 2D channel pruning methods for compressing PNN models.
The proposed {\cpt} refines the channel importance, the key factor of pruning methods, from two aspects.
Firstly, considering the point coordinates ($x$, $y$, and $z$) encode the spatial information and deeply affects feature extraction procedures in PNN layers,
we determine the channel importance by evaluating the correlation between the feature map and its corresponding point coordinates by introducing a coordinate-enhancement module.
Secondly, calculating channel importance in channel pruning is data-driven and sensitive to the input, and the intrinsic sampling steps in PNN naturally makes pruning methods unstable.
To settle this problem, we make full use of the discarded points in the sampling process via a knowledge recycling module to supplement the evaluation of channel importance.
This reduces the data sampling bias impact on the channel importance calculation and increases the robustness of the pruning results.
Notably, both the coordinates and recycled points in {\cpt} do not participate in network training (with back-propagation) but only assist channel importance calculation in the reasoning phase.
Thus, {\cpt} does not increase any computational cost of the pruned network.
The contributions of this paper are as follows:
\begin{itemize}[itemsep=1pt,topsep=1pt,parsep=1pt,leftmargin=10pt] 
\item We systematically consider the characteristics of PNNs and propose a channel pruning plug-in named CP$^3$ to enhance 2D CNN channel pruning approaches on 3D PNNs. To the best of our knowledge, CP$^3$ is the first method to export existing 2D pruning methods to PNNs.

\item We propose a coordinate-enhanced channel importance score to guide point clouds network pruning, by evaluating the correlation between feature maps and corresponding point coordinates.

\item We design a knowledge recycling pruning scheme that increases the robustness of the pruning procedure, using the discarded points to improve the channel importance evaluation.

\item We show that using CP$^3$ is consistently superior to directly transplanting 2D pruning methods to PNNs by extensive experiments on three 3D tasks and five datasets with different PNN models and pruning baselines.
\end{itemize}

\section{Related Work}

\subsection{2D Channel Pruning}
Channel pruning (a.k.a., filter pruning) methods reduce the redundant filters while maintaining the original structure of CNNs and is friendly to prevailing inference acceleration engines such as TensorFlow-Lite~(TFLite)~\cite{TensorFlow-Lite} and Mobile Neural Network~(MNN)~\cite{Ali-MNN}.
Mainstream channel pruning methods~\cite{ding2019centripetal,meng2020pruning,ding2021resrep,guan2022dais} usually first evaluate the importance of channels by certain metrics and then prune (i.e., remove) the less important channels.
Early work~\cite{li2016pruning} uses the $l_1$ norm of filters as importance score for channel pruning.
Afterwards, learning parameters, such as the scaling factor $\gamma$ in the batch norm layer~\cite{liu2017learning} and the reconstruction error in the final network layer~\cite{yu2018nisp}, are considered as the importance scores for channel selection.
The importance sampling distribution of channels~\cite{DBLP:conf/iclr/LiebenweinBLFR20} is also used for pruning.
Recent works~\cite{DBLP:conf/cvpr/HouQSMYXC00K22,sui2021chip} measure the correlation of multiple feature maps to determine the importance score of the filter for pruning.
HRank~\cite{lin2020hrank} proposes a method for pruning filters based on the theory that low-rank feature maps contain less information.
\cite{ye2020accelerating} leverages the statistical distribution of activation gradient and takes the smaller gradient as low importance score for pruning.
\cite{DBLP:conf/ijcai/WangFGCH19} calculates  the average importance of both the input feature maps and their corresponding output feature maps to determine the overall importance.
\cite{DBLP:conf/icml/000100CHL00L21, han2020model} compress CNNs from multiple dimensions
While most channel pruning methods are designed for and tested on 2D CNNs, our {\cpt} can work in tandem with existing pruners for 3D point-based networks.

\subsection{Point-based Networks for Point Cloud Data}
Point-based Neural Networks~(PNNs) directly process point cloud data with a flexible range of receptive field,
have no positioning information loss,
and thus keep more accurate spatial information.
As a pioneer work, PointNet~\cite{qi2017pointnet} learns the spatial encoding directly from the input point clouds and uses the characteristics of all points to obtain the global representations.
PointNet++~\cite{qi2017pointnet++} further proposes a multi-level feature extraction structure to extract local and global features more effectively.
KPConv~\cite{2019KPConv} proposes a new point convolution operation to learn local movements applied to kernel points.
ASSANet~\cite{2021ASSANet} proposes a separable set abstraction module that decomposes the normal SA module in PointNet++ into two separate learning phases for channel and space.
PointMLP~\cite{DBLP:conf/iclr/MaQYR022} uses residual point blocks to extract local features, transforms local points using geometric affine modules, and extracts geometric features before and after the aggregation operation.
PointNeXt~\cite{qian2022pointnext} uses inverted residual bottleneck and separable multilayer perceptrons to achieve more efficient model scaling.
Besides classification,
PNNs also serve as backbones for other 3D tasks.
VoteNet~\cite{qi2019deep} effectively improves the 3D object detection accuracy through the Hough voting mechanism~\cite{DBLP:journals/ijcv/LeibeLS08}.
PointTransformer~\cite{DBLP:conf/iccv/ZhaoJJTK21} designs models improving prior work across domains and tasks.
GroupFree3D~\cite{liu2021group1} uses the attention mechanism to automatically learn the contribution of each point to the object.
In this paper, we show that {\cpt} can be widely applied to point-based networks on a variety of point cloud benchmarks and representative original networks.

\section{Methodology}
\begin{figure*}[t]
    \centering
    \includegraphics[width=0.99\textwidth]{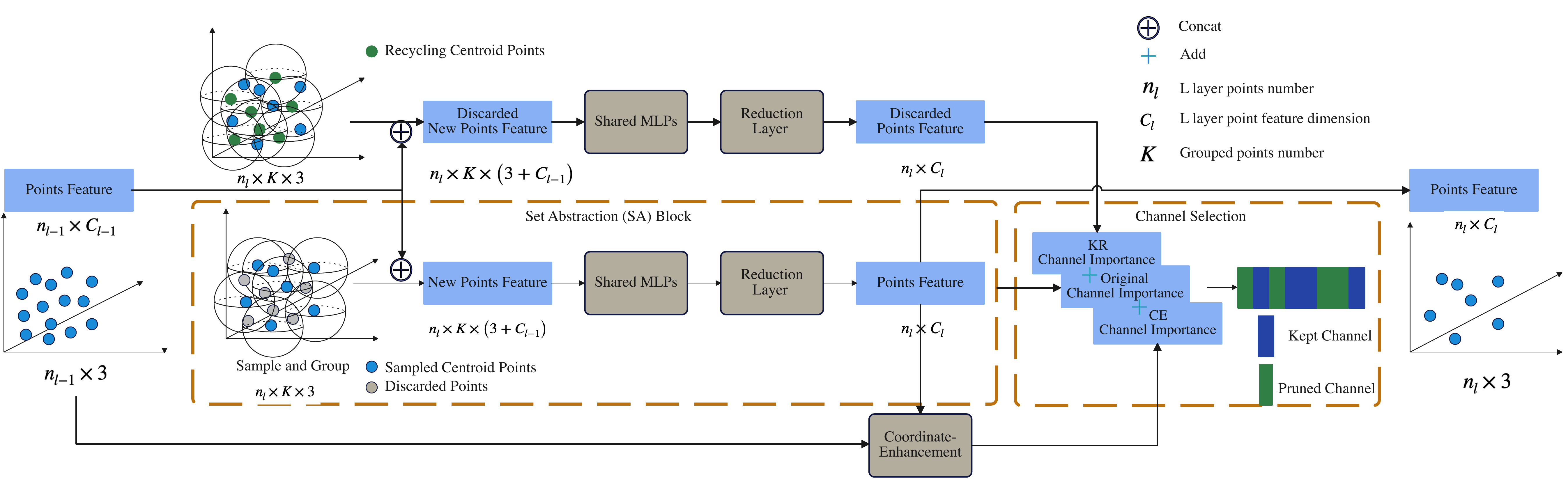}
    \vspace{-5pt}
    \caption{
    The framework of~{\cpt}. The figure shows the specific pruning process of one of the SA blocks.
    Whether a channel in a PNN is pruned is determined by three parts:
    1) Original channel importance: obtained from the original CNNs channel pruning method~(e.g., HRank \cite{lin2020hrank}, CHIP \cite{sui2021chip}).
    2) Discarded channel importance: obtained from the Knowledge-Recycling module by leveraging the discarded points in the network to supplement the channel importance evaluation of the corresponding points and improve the robustness of the channel selection.
    3) CE (Coordinate-Enhanced) channel importance: obtained from calculating the correlation between the feature map and its corresponding points coordinates to guide point clouds network pruning.
}
    \label{fig_my_graph}
\vspace{-5pt}
\end{figure*}

Although point-based networks are similar to CNN in concrete realization, they have fundamental differences in data representation and network architecture design.
To extend the success of CNN pruning to PNN, two modules are proposed in~{\cpt} taking advantage from the dimensional information and discarded points:
1) coordinate-enhancement~(CE) module, which produces a coordinate-enhanced score to estimate the channel importance by combining dimensional and feature information,
and 2) knowledge recycling module reusing the discarded points to improve the channel importance evaluation criteria and increase the robustness.

\subsection{Formulations and Motivation}
\label{sec3.1}
\para{Point-based networks}
PNN is a unified architecture that directly takes point clouds as input.
It builds hierarchical groups of points and progressively abstracts larger local regions along the hierarchy.
PNN is structurally composed by a number of set abstraction~(SA) blocks.
Each SA block consists of
1) a sampling layer iteratively samples the farthest point to choose a subset of points from input points,
2) a group layer gathers neighbors of centroid points to a local region,
3) a set of shared Multi-Layer Perceptrons~(MLPs) to extract features,
and 4) a reduction layer to aggregate features in the neighbors.
Formally speaking, a SA block takes an $n_{i-1} \times (d+c_{i-1})$ matrix as input that is from $n_{i-1}$ points with $d$-dim coordinates and $c_{i-1}$-dim point feature. It outputs an $n_i \times (d + c_i)$ matrix of $n_i$ subsampled points with $d$-dimensional coordinates (i.e., $d=3$) and new $c_i$-dimensional feature vectors summarizing local context.
The SA block is formulated as:
\begin{equation}
\label{point_based_eq}
\mathbf{F}_i^{l+1}=\mathcal{R}\left\{h_{\Theta}\left(\left[\mathbf{F}_j^l ; \mathbf{x}_j^l-\mathbf{x}_i^l\right]\right)\right\},
\end{equation}
where $h_{\Theta}$ is MLPs to extract grouped points feature, $\mathcal{R}$ is the reduction layer (e.g. max-pooling)  to aggregate features in the neighbors $\{j:(i, j) \in \mathcal{N}\}$, $F_j^l$ is the features of neighbor $j$ in the $l$-th layer, $x_i^l$ and $x_j^l$ are input points coordinates and coordinates of neighbor $j$ in the $l$-th layer.

\para{Channel pruning}
Assume a pre-trained PNN model has a set of $K$ convolutional layers, and $\mathcal{A}^l$ is the $l$-th convolution layer. The parameters in $\mathcal{A}^l$ can be represented as a set of filters $\mathcal{W}_{\mathcal{A}^l}=\left\{\mathbf{w}_1^l, \mathbf{w}_2^l, \ldots, \mathbf{w}_{c_l}^l\right\} \in \mathbb{R}^{(d + c_l) \times (d + c_{l-1}) \times k_l \times k_l}$, where $j$-th filter is $\mathbf{w}_j^l \in \mathbb{R}^{(d + c_{l-1}) \times k_l \times k_i}$. $(d +c_l)$ represents the number of filters in $\mathcal{A}^l$ and $k_l$ denotes the kernel size. The outputs of filter, i.e., feature map, are denoted as $\mathcal{F}^l=\left\{\mathbf{f}_1^l, \mathbf{f}_2^l, \ldots, \mathbf{f}_{n_i}^l\right\} \in \mathbb{R}^{n_i \times (d+c_i)}$.
Channel pruning aims to identify and remove the less importance filter from the original networks. In general, channel pruning can be formulated as the following optimization problem:
\begin{equation}
\label{filter_pruning}
\min _{\delta_{i j}} \sum_{i=1}^K \sum_{j=1}^{n_i} \delta_{i j} \mathcal{L}\left(\mathbf{w}_j^i\right) \text {, s.t. }\sum_{j=1}^{n_i} \delta_{i j}=k_l,
\end{equation}
where $\delta_{i j}$ is an indicator which is 1 if $\mathbf{w}_j^i$ is to be pruned or 0 if  $\mathbf{w}_j^i$ is to be kept, $\mathcal{L}(\cdot)$ measures the importance of a filter and $k_l$ is the kept filter number.

\para{Robust importance metric for channel pruning}
The metrics for evaluating the importance of filters is critical. Existing CNN pruning methods design a variety of $\mathcal{L}(\cdot)$ on the filters.
Consider the feature maps, contain rich and important information of both filter and input data, approaches using feature information have become popular and achieved state-of-the-art performance for channel pruning.
The results of the feature maps may vary depending on the variability of the input data. Therefore, when the importance of one filter solely depends on the information represented by its own generated feature map, the measurement of the importance may be unstable and sensitive to the slight change of input data.
So we have taken into account the characteristics of point clouds data and point-based networks architecture to improve the robustness of channel importance in point-based networks.
On the one hand, we propose a coordinate-enhancement module by evaluating the correlation between the feature map and its corresponding points coordinates to guide point clouds network pruning, which will be described in Sec. \ref{sec3.2}.  On the other hand, we design a knowledge recycling pruning schema, using discarded points to improve the channel importance evaluation criteria and increase the robustness of the pruning module, which will be described in detail in Sec.~\ref{sec3.3}.

\subsection{Coordinate-Enhanced Channel Importance} \label{sec3.2}
Dimensional information is critical in PNNs. The dimensional information (i.e., coordinates of the points) are usually adopted as input for feature extraction. Namely, the input and output of each SA block are concatenated with the coordinates of the points.
Meanwhile, the intermediate feature maps reflect not only the information of the original input data but also the corresponding channel information.
Therefore, the importance of the channel can be obtained from the feature maps, i.e., the importance of the corresponding channel.
The dimensional information is crucial in point-based tasks and should be considered as part of importance metric.
Thus the critical problem falls in designing a function that can well reflect the dimensional information richness of feature maps.
The feature map, obtained by encoding points spatial $x$, $y$, and $z$ coordinates, should be closely related to the original corresponding points coordinates.
Therefore, we use the correlation between the current feature map and the corresponding input points coordinates to determine the importance of the filter.
The designed Coordinate-Enhancement (CE) module based on Eq.~\eqref{filter_pruning}:
\begin{equation}
\label{filter_pruning_c}
\min _{\delta_{i j}} \sum_{i=1}^K \sum_{j=1}^{n_i} \delta_{i j} \mathcal{L}_c\left(\mathbf{F}_j^i\right) \text {, s.t. }\sum_{j=1}^{n_i} \delta_{i j}=k_l,
\end{equation}
where $\delta_{i j}$ is an indicator which is 1 if $\mathbf{w}_j^i$ is to be pruned or 0 if  $\mathbf{w}_j^i$ is to be kept and $k_l$ is the kept channel number. $\mathcal{L}_c(\cdot)$ measures the importance of a channel from take account of the relationship between feature map and points coordinates which can be formulated as:
\begin{equation}
\label{ce_pruned}
\mathcal{L}_c\left(F_j^i\right)=\mathcal {M}\{\operatorname {CE}\left(F_j^i, x^i\right)\},
\end{equation}
where $\operatorname {CE}$ obtains the coordinate-enhanced score by calculating correlation of each channel in the feature map with the original coordinates, and $\mathcal {M}$ takes the maximum value.
Hence, higher coordinate-enhanced score (i.e., $\mathcal{L}_c$) serve as a reliable measurement for information richness.

\begin{figure}[t]
  \centering
  \includegraphics[width=.8\columnwidth]{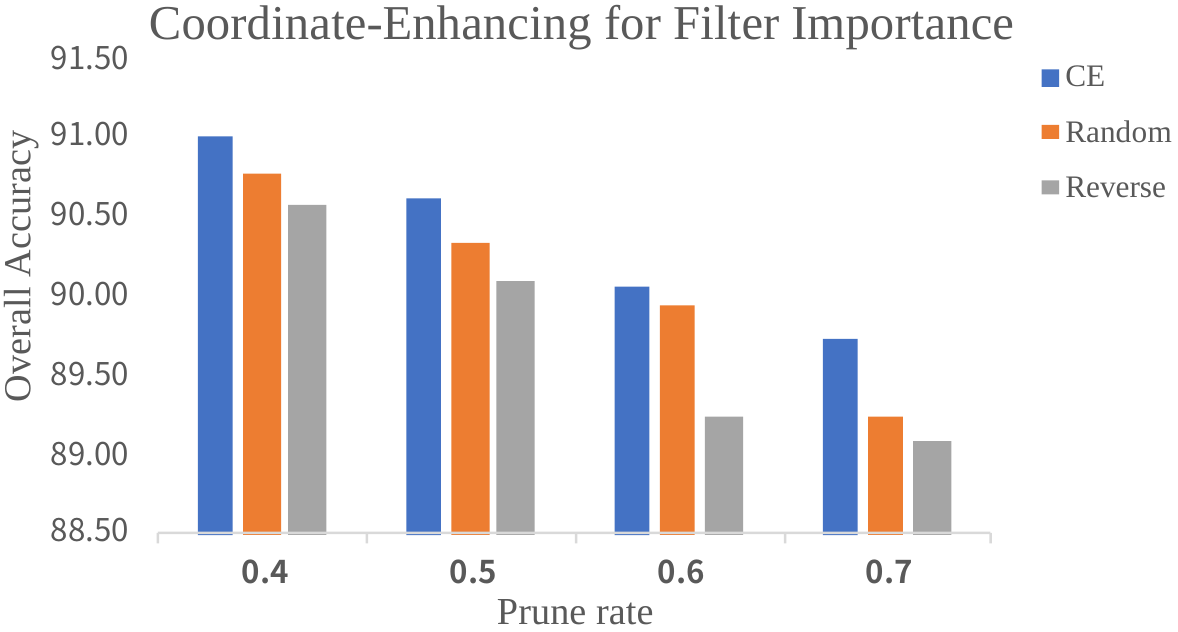}
  \vspace{-5pt}
  \caption{Comparisons on the pruned model accuracy with different pruning metrics with CE scores. Results are on ModelNet40 with PointNeXt-S~(C=64).}
  \label{cd_filter}
  \vspace{-5pt}
\end{figure}

We evaluated the effectiveness of the CE module on ModelNet40 with PointNeXt-S~(C=64).
To demonstrate the experiment's validity, we compared the overall accuracy (OA) for pruning rates of 40\%, 50\%, 60\%, and 70\%, respectively.
Three sets of experiments are carried out. First, the `CE' group selected filters in order of value from the CE module. Secondly, the `Random' group is used to randomly select filters for pruning, and finally, the `Reverse' is used to select filters according to the coordinate-enhanced score from low to high. As shown in Fig.~\ref{cd_filter}, the channel with a higher coordinate-enhanced score has higher accuracy, which means that the channel with a higher coordinate-enhanced score has higher importance and should be retained in the pruning process.
\subsection{Knowledge Recycling}\label{sec3.3}

Sec.~\ref{sec3.2} shows that feature maps can reflect the importance of the corresponding channels.
Another problem is that the importance determination may be unstable and sensitive to small changes in the input data as the feature maps are highly related to the samples of input data.
Therefore, we aim to reduce the impact of such data variation.
Through the analysis of the PNNs in Sec.~\ref{sec3.1}, we found that some points are discarded to obtain hierarchical points set feature.
These discarded points are informative as well since the sampling mechanism are highly random and can be leveraged to reduce the impact of data variation.
The Knowledge Recycling~(KR) module is proposed to reuse the discarded points to improve the robustness of channel pruning.
\begin{table*}[ht]
\caption{Comparisons of classification on the ModelNet40~\cite{DBLP:conf/cvpr/WuSKYZTX15} test set with PointNet++~\cite{qi2017pointnet++}, PointNeXt-S (C=32)~\cite{qian2022pointnext}, and PointNeXt-S (C=64).
For PointNeXt-S (C=32), we report the baseline results from the original paper.
For PointNet++ and PointNeXt-S (C=64), we report the baseline results obtained by OpenPoints~\cite{qian2022pointnext} re-implementations trained with the improved strategies.}
\vspace{-5pt}
\label{modelnet40_sum_label}
\resizebox{\linewidth}{!}{
\begin{tabular}{l|cccc|cccc|cccc}
\toprule
\multirow{2.5}{*}{\textbf{Method}}
            & \multicolumn{4}{c|}{\textbf{PointNet++}}        & \multicolumn{4}{c|}{\textbf{PointNeXt-S (C=32)}} & \multicolumn{4}{c}{\textbf{PointNeXt-S (C=64)}} \\ \cmidrule(lr){2-5} \cmidrule(lr){6-9} \cmidrule(lr){10-13}
            & \textbf{OA}    & \textbf{mAcc}  & \textbf{Params. (M)} & \textbf{GFLOPs ($\downarrow \%$)}       & \textbf{OA}     & \textbf{mAcc}  & \textbf{Params. (M)} & \textbf{GFLOPs ($\downarrow \%$)} & \textbf{OA}     & \textbf{mAcc}  & \textbf{Params. (M)} & \textbf{GFLOPs ($\downarrow \%$)}\\ \midrule
Baseline    & 92.80  & 89.90  & 1.47    & 1.71 (--)     & 92.99  & 89.6  & 1.37    & 1.64 (--)    & 93.44  & 91.05 & 4.52    & 6.49 (--)    \\
\cmidrule(lr){1-13}
HRank       & 92.59 & 89.83 & 0.86    & 0.77 (55.0)  & 92.87  & 89.97 & 0.74    & 0.71 (56.7) & 92.23  & 89.81 & 2.12    & 2.69 (56.7) \\
HRank +{\cpt}  & 92.95 & 89.91 & 0.84    & 0.75 (56.1)  & 93.23  & 90.56 & 0.71    & 0.67 (59.1) & 93.52  & 90.33 & 2.01    & 2.58 (59.1) \\
HRank        & 91.79 & 88.82 & 0.59    & 0.42 (75.4)  & 92.63  & 89.12 & 0.50    & 0.39 (76.2) & 92.71  & 90.45 & 1.33    & 1.56 (76.2) \\
HRank +{\cpt}  & 92.54 & 88.52 & 0.57    & 0.39 (77.2)  & 93.03  & 90.92 & 0.49    & 0.38 (76.8) & 93.07  & 90.55 & 1.28    & 1.50 (76.8) \\
HRank        & 91.34 & 88.18 & 0.36    & 0.15 (91.2)  & 92.73  & 89.98 & 0.43    & 0.29 (82.3) & 92.83  & 89.99 & 0.74    & 0.71 (82.3) \\
HRank +{\cpt}  & 91.71 & 88.68 & 0.34    & 0.13 (92.4)  & 92.99  & 90.11 & 0.40    & 0.27 (83.5) & 93.11  & 90.52 & 0.71    & 0.67 (83.5) \\ \cmidrule(lr){1-13}
ResRep      & 92.71 & 90.39 & 0.85    & 0.73 (57.3)  & 92.83  & 90.15 & 0.81    & 0.69 (57.9) & 91.61  & 88.42 & 2.08    & 2.34 (57.9) \\
ResRep+{\cpt} & 93.27 & 90.48 & 0.82    & 0.70 (59.1)  & 93.35  & 90.93 & 0.79    & 0.67 (59.1) & 92.93  & 90.66 & 1.89    & 2.02 (59.1) \\
ResRep      & 92.50  & 89.25 & 0.57    & 0.41 (76.0)  & 92.64  & 90.01 & 0.51    & 0.40 (75.6) & 91.67  & 89.27 & 1.13    & 1.92 (75.6) \\
ResRep+{\cpt} & 92.46 & 89.43 & 0.54    & 0.40 (76.6)  & 93.41  & 90.87 & 0.49    & 0.38 (76.8) & 93.11  & 90.82 & 1.02    & 1.89 (76.8) \\
ResRep      & 92.11 & 89.00    & 0.55    & 0.24 (86.0)) & 92.30  & 89.31 & 0.34    & 0.21 (87.2) & 89.54  & 88.54 & 0.72    & 0.69 (87.2) \\
ResRep+{\cpt} & 92.48 & 89.21 & 0.58    & 0.21 (87.7)  & 92.95  & 90.70  & 0.33    & 0.18 (89.0) & 91.02  & 89.82 & 0.69    & 0.65 (89.0) \\ \cmidrule(lr){1-13}
CHIP        & 92.79 & 89.23 & 0.82    & 0.73 (57.3)  & 93.11  & 90.27 & 0.71    & 0.67 (59.1) & 93.03  & 90.60  & 1.48    & 1.44 (59.1) \\
CHIP+{\cpt}   & 92.99 & 90.66 & 0.81    & 0.70 (59.1)  & 93.35  & 90.80  & 0.69    & 0.65 (60.4) & 93.35  & 91.11 & 1.45    & 1.40 (60.4) \\
CHIP        & 92.45 & 89.19 & 0.57    & 0.39 (77.2)  & 92.71  & 89.90  & 0.49    & 0.38 (76.8) & 92.79  & 90.39 & 0.92    & 0.76 (76.8) \\
CHIP+{\cpt}   & 92.91 & 89.65 & 0.54    & 0.35 (79.5)  & 93.03  & 90.51 & 0.48    & 0.37 (77.4) & 93.23  & 90.30  & 0.89    & 0.74 (77.4) \\
CHIP        & 92.26 & 89.56 & 0.36    & 0.15 (91.2)  & 92.42  & 89.13 & 0.32    & 0.16 (90.2) & 90.83  & 88.70  & 0.65    & 0.46 (90.2) \\
CHIP+{\cpt}   & 92.71 & 90.41 & 0.34    & 0.13 (92.4)  & 92.50  & 90.35 & 0.30    & 0.14 (91.5) & 92.87  & 90.25 & 0.63    & 0.44 (91.5) \\ \bottomrule
\end{tabular}}
\end{table*}

For those centroids that are computed in $(l-1)$-th layer but discarded in $l$-th layer due to sampling, which are equivalent to the sampled points.
Therefore, the discarded centroids $x_{dis}$ are feeded into $l$-th convolutional layer to generate the feature map $\mathbf{f_{dis}}^l$, and $\mathcal{L}_k\left(F_{dis_j}^i\right)$ is taken in use for the evaluation of channel importance.

We calculate the relevant feature maps from the network parameters trained from the sampled points and use them as part of the importance calculation for the current SA layer channels.
Specifically, for each layer in the SA module, we obtain the features of the discard points by Eq. 1:
$$
\mathcal{F}_{dis}^l=\left\{\mathbf{f_{dis}}_1^l, \mathbf{f_{dis}}_2^l, \ldots, \mathbf{f_{dis}}_{n_i}^l\right\} \in \mathbb{R}^{n_i \times (3+c_i)},
$$
where $n_i$ is sampled points number, $c_i$ is the points feature dimension.
So the supplement importance is:
\begin{equation}
\label{kr_prund}
\begin{aligned}
\mathcal{L}_k\left(F_j^i\right)&=\mathcal{L}_k\left(F_{dis_j}^i\right)\\
                     &=\mathcal {M}\{\operatorname {CE}\left(F_j^i, x_{dis}^i\right)\},
\end{aligned}
\end{equation}
where $x_{dis}^i$ are discard points for recycling.

It should be noted that the KR module only needs to calculate $\mathcal{F}_{dis}^l$ from the parameters trained by the sampled points and does not incur much additional overhead.

\subsection{Using {\cpt} in Pruning Methods}

The overall {\cpt} improve the existing CNNs pruning methods by considering the input data of PNNs and the PNN structure in Sec.~\ref{sec3.2} and Sec.~\ref{sec3.3}, respectively.
In fact, {\cpt} can complement the existing pruning methods, i.e., as a plug-in to the existing pruning methods, to improve the pruning performance on PNNs.
Specifically, combining Eq.~\eqref{ce_pruned} and Eq.~\eqref{kr_prund}, we obtain the final pruning formula according to Eq.~\eqref{filter_pruning}:
\begin{equation}
\begin{aligned}
\min _{\delta_{i j}} \sum_{i=1}^K \sum_{j=1}^{n_i} \delta_{i j} (&\mathcal{L}\left(\mathbf{w}_j^i\right)+\mathcal{L}_c\left(F_j^i\right)+\mathcal{L}_k\left(F_j^i\right)), \\
&\text {s.t. }\sum_{j=1}^{n_i} \delta_{i j}=k_l,
\end{aligned}
\end{equation}
where $\delta_{i j}$ is an indicator which is 1 if $\mathbf{w}_j^i$ is to be pruned or 0 if  $\mathbf{w}_j^i$ is to be kept, $\mathcal{L}(\cdot)$ is original CNNs pruning method measure importance of a channel, $\mathcal{L}_c$ and $\mathcal{L}_k$ are coordinate-enhanced score and knowledge-recycling score, $k_l$ is the kept channel number.

\section{Experiments}
\subsection{Experimental Settings}
\para{Baseline models and datasets}
To demonstrate the effectiveness and generality of the proposed {\cpt}, we tested it on three different 3D tasks and five datasets with various PNNs and three recent advanced channel pruning methods.
The evaluated pruning methods include HRank~(\textit{2020})~\cite{lin2020hrank}, ResRep~(\textit{2021})~\cite{ding2021resrep}, and CHIP~(\textit{2021})~\cite{sui2021chip}.
For the \textit{classification task}, we chose the classical PointNet++~\cite{qi2017pointnet++} and PointNeXt~\cite{qian2022pointnext} models as the original networks and conducted experiments on ModelNet40~\cite{DBLP:conf/cvpr/WuSKYZTX15} and ScanObjectNN~\cite{DBLP:conf/iccv/UyPHNY19}. Specifically, for PointNeXt-S we tested two settings with widths of 32 and 64.
For the \textit{segmentation} task, we conducted experiments on S3DIS~\cite{DBLP:conf/cvpr/ArmeniSZJBFS16} with PointNeXt-B and PointNeXt-L~\cite{qian2022pointnext} as the original PNNs.
For the \textit{object detection} task, we pruned two point-based detectors (VoteNet~\cite{qi2019deep} and GroupFree3D~\cite{liu2021group}) on \mbox{SUN RGB-D}~\cite{DBLP:conf/cvpr/SongLX15} and ScanNetV2~\cite{DBLP:conf/cvpr/DaiCSHFN17}.

\para{Implementation details}
We conducted the classification and segmentation experiments with OpenPoints~\cite{qian2022pointnext} and the object detection experiments with MMdetection3D~\cite{mmdet3d2020}, all on NVIDIA P100 GPUs.
For a fair comparison, we used the same hyperparameter settings for each group of experiments.
We either 1) measured the parameter/FLOP reductions of the pruned networks with similar performance or 2) measured the performance of the pruned networks with a similar amount of parameter/FLOP reductions.
For all experiments, we reported the number of FLOPs (`GFLOPs') and parameters (`Params.'), as well as task-specific metrics to be described in each experiment.
More experimental results are available in the supplementary.

\subsection{Results on Classification}

\para{ModelNet40}
ModelNet40~\cite{DBLP:conf/cvpr/WuSKYZTX15} contains 9843 training and 2468 testing meshed CAD models belonging to 40 categories.
Following the standard practice~\cite{qi2017pointnet++}, we report the class-average accuracy~(mAcc) and the overall accuracy~(OA) on the testing set.
We compared the pruned networks directly by HRank, ResRep, and CHIP and those pruned by the three pruning methods with {\cpt}.
As shown in Tab.~\ref{modelnet40_sum_label},
our {\cpt} improved the performance of existing CNNs pruning methods for different PNNs with various pruning rates.
{\cpt} had higher accuracy scores with a similar (and mostly higher) pruning rates.
Notably, with the pruning rate of ~58\%, {\cpt} usually produced compact PNNs with even better accuracy scores than the original PNNs, which is difficult for pruning methods without {\cpt}.

\begin{table*}[ht]
\caption{Comparisons of classification on the ScanObjectNN \cite{DBLP:conf/iccv/UyPHNY19} test set with PointNet++~\cite{qi2017pointnet++}, PointNeXt-S (C=32)~\cite{qian2022pointnext}, and PointNeXt-S (C=64).
For PointNeXt-S (C=32), we report the baseline results from the original paper.
For PointNet++ and PointNeXt-S (C=64), we report the baseline results obtained by OpenPoints~\cite{qian2022pointnext} re-implementations trained with the improved strategies.}
\vspace{-5pt}
\label{scanobject_sum_label}
\resizebox{\linewidth}{!}{
\begin{tabular}{l|cccc|cccc|cccc}
\toprule
\multirow{2.5}{*}{\textbf{Method}}
            & \multicolumn{4}{c|}{\textbf{PointNet++}}        & \multicolumn{4}{c|}{\textbf{PointNeXt-S (C=32)}} & \multicolumn{4}{c}{\textbf{PointNeXt-S (C=64)}} \\
            \cmidrule(lr){2-5} \cmidrule(lr){6-9} \cmidrule(lr){10-13}
            & \textbf{OA}    & \textbf{mAcc}  & \textbf{Params. (M)} & \textbf{GFLOPs ($\downarrow \%$)}       & \textbf{OA}     & \textbf{mAcc}  & \textbf{Params. (M)} & \textbf{GFLOPs ($\downarrow \%$)} & \textbf{OA}     & \textbf{mAcc}  & \textbf{Params. (M)} & \textbf{GFLOPs ($\downarrow \%$)}\\ \midrule
\multicolumn{1}{l|}{Baseline}    & 86.20  & 84.40  & 1.47       & 1.71 (--)    & 87.40  & 85.39 & 1.37       & 1.64 (--)    & 88.20  & 86.84 & 4.52       & 6.49 (--)    \\  \cmidrule(lr){1-13}
\multicolumn{1}{l|}{HRank}       & 85.01 & 83.33 & 0.82       & 0.73 (57.3) & 87.02 & 85.85 & 0.72       & 0.69 (57.9) & 87.51 & 85.48 & 2.22       & 2.91 (55.2) \\
\multicolumn{1}{l|}{HRank+{\cpt}}  & 86.15 & 84.40 & 0.80       & 0.70 (59.1) & 87.47 & 86.21 & 0.70       & 0.67 (59.1) & 87.95 & 86.02 & 2.16       & 2.82 (56.5) \\
\multicolumn{1}{l|}{HRank}       & 84.94 & 83.43 & 0.59       & 0.43 (74.9) & 84.79 & 81.93 & 0.50       & 0.39 (76.2) & 84.66 & 81.00 & 1.33       & 1.56 (76.0) \\
\multicolumn{1}{l|}{HRank+{\cpt}}  & 86.08 & 84.51 & 0.59       & 0.41 (76.0) & 86.40 & 83.94 & 0.48       & 0.37 (77.4) & 86.43 & 84.94 & 1.28       & 1.50 (76.9) \\
\multicolumn{1}{l|}{HRank}       & 84.03 & 82.16 & 0.37       & 0.16 (90.6) & 81.33 & 78.32 & 0.32       & 0.17 (89.6) & 85.39 & 83.83 & 0.73       & 0.71 (89.1) \\
\multicolumn{1}{l|}{HRank+{\cpt}}  & 84.62 & 82.29 & 0.36       & 0.15 (91.2) & 84.21 & 82.13 & 0.31       & 0.16 (90.2) & 86.26 & 84.36 & 0.70       & 0.67 (89.7) \\ \cmidrule(lr){1-13}
\multicolumn{1}{l|}{ResRep}      & 86.77 & 84.78 & 0.82       & 0.76 (55.6) & 86.32 & 84.45 & 0.69       & 0.65 (60.4) & 86.58 & 84.02 & 2.12       & 2.82 (56.5) \\
\multicolumn{1}{l|}{ResRep+{\cpt}} & 87.11 & 85.50 & 0.82       & 0.71 (58.5) & 86.66 & 84.75 & 0.67       & 0.64 (61.0) & 88.52 & 86.00 & 2.03       & 2.74 (57.8) \\
\multicolumn{1}{l|}{ResRep}      & 83.99 & 82.34 & 0.60       & 0.44 (74.3) & 85.28 & 83.27 & 0.48       & 0.38 (76.8) & 85.22 & 82.03 & 1.39       & 1.65 (74.6) \\
\multicolumn{1}{l|}{ResRep+{\cpt}} & 84.32 & 83.68 & 0.59       & 0.43 (74.9) & 86.22 & 84.32 & 0.47       & 0.37 (77.4) & 86.73 & 84.23 & 1.32       & 1.54 (76.3) \\
\multicolumn{1}{l|}{ResRep}      & 83.79 & 81.91 & 0.40       & 0.29 (83.0) & 83.66 & 81.66 & 0.41       & 0.25 (84.8) & 85.02 & 81.27 & 0.60       & 0.68 (89.5) \\
\multicolumn{1}{l|}{ResRep+{\cpt}} & 84.80 & 82.83 & 0.40       & 0.26 (84.8) & 85.01 & 83.03 & 0.40       & 0.23 (86.0) & 86.13 & 83.40 & 0.58       & 0.52 (92.0) \\ \cmidrule(lr){1-13}
\multicolumn{1}{l|}{CHIP}        & 86.14 & 84.98 & 0.82       & 0.73 (57.3) & 87.13 & 85.09 & 0.70       & 0.67 (59.1) & 88.45 & 87.40 & 2.11       & 2.74 (57.8) \\
\multicolumn{1}{l|}{CHIP+{\cpt}}   & 86.25 & 84.68 & 0.80       & 0.70 (59.1) & 87.54 & 85.67 & 0.69       & 0.65 (60.4) & 88.58 & 86.45 & 2.05       & 2.65 (59.2) \\
\multicolumn{1}{l|}{CHIP}        & 84.45 & 82.25 & 0.59       & 0.42 (75.4) & 85.28 & 83.28 & 0.50       & 0.4 (75.6)  & 86.68 & 85.12 & 1.37       & 1.64 (74.7) \\
\multicolumn{1}{l|}{CHIP+{\cpt}}   & 85.59 & 84.42 & 0.57       & 0.40 (76.6) & 86.29 & 83.99 & 0.49       & 0.38 (76.8) & 87.79 & 86.81 & 1.32       & 1.56 (76.0) \\
\multicolumn{1}{l|}{CHIP}        & 83.27 & 81.15 & 0.36       & 0.15 (91.2) & 81.37 & 78.99 & 0.34       & 0.19 (88.4) & 83.90  & 81.83 & 0.44       & 0.32 (95.1) \\
\multicolumn{1}{l|}{CHIP+{\cpt}}   & 84.03 & 82.22 & 0.35       & 0.14 (91.8) & 82.12 & 79.41 & 0.33       & 0.18 (89.0) & 84.25 & 82.17 & 0.42       & 0.29 (95.5) \\ \bottomrule
\end{tabular}}
\end{table*}

\para{ScanObjectNN}
We also conducted experiments on the ScanObjectNN benchmark~\cite{DBLP:conf/iccv/UyPHNY19}.
ScanObjectNN contains 15000 objects categorized into 15 classes with 2902 unique object instances in the real world.
As reported in Tab.~\ref{scanobject_sum_label}, {\cpt} surpassed existing CNN pruning methods directly appled to PNNs.
For example, comparing with the baseline pruning method HRank,
{\cpt} boosts the OA score of PointNet++ by 1.14\% (85.01$\rightarrow$86.15), 1.14\% (84.94$\rightarrow$86.08), and 0.59\% (84.03$\rightarrow$84.62) for the three different pruning rates.
Similarly, {\cpt} obtains much higher OA and mAcc scores than the three baselines with different pruning rates on PointNet++ and PointNeXt-S.
With the extensive experimental results on classification tasks, we show that {\cpt} surely improved the pruned network's performance to a distinct extent compared to direct applications of 2D CNN pruning methods.

\begin{table*}[t]
\centering
\caption{Comparisons of semantic segmentation on the S3DIS dataset (evaluated in Area-5) with PointNeXt-B and PointNeXt-L~\cite{qian2022pointnext}.}
\label{seg_s3dis}
\vspace{-5pt}
\resizebox{.9\linewidth}{!}{
\begin{tabular}{l|ccccc|ccccc}
\toprule
\multirow{2.5}{*}{\textbf{Method}}
           & \multicolumn{5}{c|}{\textbf{PointNeXt-B}}                & \multicolumn{5}{c}{\textbf{PointNeXt-L}}                 \\ \cmidrule(lr){2-6} \cmidrule(lr){7-11}
           & \textbf{OA}    & \textbf{mAcc}  & \textbf{mIoU}  & \textbf{Params. (M)} & \textbf{GFLOPs ($\downarrow\%$)}     & \textbf{OA}    & \textbf{mAcc}  & \textbf{mIoU}  & \textbf{Params. (M)} & \textbf{GFLOPs ($\downarrow\%$)}     \\ \midrule
Baseline   & 89.40  & 73.90  & 67.50  & 3.83       & 8.80 (--)     & 90.10  & 75.70  & 69.30  & 7.13       & 15.24 (--)   \\  \cmidrule(lr){1-11}
HRank      & 89.04 & 72.14 & 65.66 & 1.72       & 4.01 (54.4) & 88.88 & 73.61 & 66.80  & 3.20       & 6.83 (55.2) \\
HRank+{\cpt} & 89.24 & 73.76 & 66.95 & 1.61       & 3.80 (56.8) & 89.44 & 74.27 & 67.53 & 3.00       & 6.48 (57.5) \\
HRank      & 88.81 & 72.16 & 65.58 & 0.85       & 2.04 (76.8) & 88.21 & 70.93 & 64.30  & 1.58       & 3.47 (77.2) \\
HRank+{\cpt} & 89.02 & 72.82 & 66.41 & 0.78       & 1.85 (78.9) & 88.35 & 71.26 & 64.71 & 1.44       & 3.14 (79.4) \\ \cmidrule(lr){1-11}
CHIP       & 88.89 & 73.26 & 66.57 & 1.66       & 3.93 (55.3) & 89.16 & 73.72 & 67.09 & 3.09       & 6.68 (56.2) \\
CHIP+{\cpt}  & 89.68 & 73.14 & 66.80 & 1.56       & 3.65 (58.5) & 89.47 & 74.20  & 67.28 & 2.91       & 6.22 (59.2) \\
CHIP       & 88.67 & 73.24 & 66.68 & 0.81       & 1.92 (78.2) & 88.58 & 71.58 & 65.18 & 1.50       & 3.27 (78.5) \\
CHIP+{\cpt}  & 89.81 & 73.36 & 66.95 & 0.74       & 1.79 (79.7) & 89.20 & 71.66 & 65.24 & 1.38       & 3.04 (80.1) \\ \bottomrule
\end{tabular}}
\end{table*}

\begin{table}[t]
\centering
\caption{Comparisons of object detection on the ScanNet dataset.}\label{tabl1_vsn}
\vspace{-5pt}
\resizebox{\columnwidth}{!}{
\begin{tabular}{lcccc}
\toprule
\textbf{Method} & \textbf{mAP@0.25} & \textbf{mAP@0.50} & \textbf{Params. (K)} & \textbf{GFLOPs ($\downarrow$\%)}      \\ \midrule
Baseline (VoteNet) & 62.34    & 40.82   & 641.92       & 5.78 (--)     \\ \cmidrule(lr){1-5}
ResRep   & 62.45    & 40.95   & 251.23       & 2.45 (57.6)  \\
\textbf{ResRep+{\cpt}}& \textbf{63.92} & \textbf{41.47} & \textbf{242.26} &\textbf{2.41 (58.1)}  \\ \cmidrule(lr){1-5}
ResRep   & 61.78    & 40.54   & 180.49       & 1.83 (68.3) \\
\textbf{ResRep+{\cpt}} & \textbf{62.98}& \textbf{40.94}& \textbf{160.48}& \textbf{1.78 (69.2)} \\ \bottomrule
\end{tabular}
}
\end{table}

\subsection{Results on Semantic Segmentation}
S3DIS~\cite{DBLP:conf/cvpr/ArmeniSZJBFS16} is a challenging benchmark composed of 6 large-scale indoor areas, 271 rooms, and 13 semantic categories in total.
Following a common protocol \cite{DBLP:conf/3dim/TchapmiCAGS17}, we evaluated the presented approach in Area-5, which means to test on Area-5 and to train on the rest.
For evaluation metrics, we used the mean classwise intersection over union~(mIoU), the mean classwise accuracy~(mAcc), and the overall pointwise accuracy~(OA).
As the segmentation task is relatively difficult and the segmentation network structures are relatively complex,
we pruned only the encoder part of the network and kept the original decoder part.
The results are presented in Tab.~\ref{seg_s3dis}.
As expected, the performance of the pruned networks degraded more from the original networks than those in the classification experiments, but the performance was acceptable. Meanwhile, in all cases, with our {\cpt}, PNNs have a higher accuracy at a higher pruning rate than without {\cpt}.
For example, for PointNeXt-B, comparing with directly applying CHIP, incorporating {\cpt} obtained much higher OA, mAcc and mIoU scores (1.2\% on OA and 0.3\% on mIoU).
The results about segmentation have well validated the generalization of {\cpt} to new and difficult tasks.

\begin{table}[t]
\centering
\caption{Comparisons of object detection on SUN RGB-D.}\label{table2_vsun}
\vspace{-5pt}
\resizebox{\columnwidth}{!}{
\begin{tabular}{lcccc}
\toprule
\textbf{Method} & \textbf{mAP@0.25} & \textbf{mAP@0.50} & \textbf{Params. (K)} & \textbf{GFLOPs ($\downarrow$\%)}      \\ \midrule
Baseline (VoteNet)        & 59.78             & 35.77             & 641.92               & 5.78 (--)              \\  \cmidrule(lr){1-5}
ResRep          & 59.37             & 36.80             & 179.91               & 2.42 (58.13)          \\
\textbf{ResRep+{\cpt}}   & \textbf{60.10}    & \textbf{37.37}    & \textbf{172.35}      & \textbf{2.20 (61.93)} \\  \cmidrule(lr){1-5}
ResRep          & 59.01             & 35.91             & 135.13               & 1.84 (68.17)          \\
\textbf{ResRep+{\cpt}}   & \textbf{59.18}    & \textbf{36.24}    & \textbf{129.64}      & \textbf{1.83 (68.34)} \\ \bottomrule
\end{tabular}
}
\end{table}

\subsection{Results on 3D Object Detection}
\subsubsection{Evaluation and Comparison of VoteNet}

Tabs.~\ref{tabl1_vsn}~and~\ref{table2_vsun} show the results of the pruned VoteNet models on the ScanNetV2 and SUN RGB-D datasets, respectively.
We evaluated the performance of our proposed method in terms of the mean average precision at IOU threshods of 0.25 and 0.50 (mAP@25 and mAP@50).

\para{ScanNetV2}
ScanNetV2~\cite{DBLP:conf/cvpr/DaiCSHFN17} is a richly annotated dataset of 3D reconstructed meshes of indoor scenes.
It contains about 1200 training examples collected from hundreds of different rooms and is annotated with semantic and instance segmentation for 18 object categories.
Tab.~\ref{tabl1_vsn} shows the results of directly applying ResRep and with {\cpt}.
As can be seen from the table, the accuracy of the 2D method directly applied to the 3D network decreased by a flops drop of about 60\%, while our method achieves 1.58\% and 0.65\% improvements at mAP@0.25 and mAP@0.5 with a drop rate of 58.13\% FLOPS.
When FLOPs drop to about 70\%, the accuracy of the direct porting CNNs pruning method works poorly, while the improvement of mAP@0.25 and mAP@0.5 of our method is 0.64\% and 0.12\%, respectively.

\para{SUN RGB-D}
The SUN RGB-D dataset~\cite{DBLP:conf/cvpr/SongLX15} consists of 10355 single-view indoor RGB-D images annotated with over 64000 3D bounding boxes and semantic labels for 37 categories.
We conducted experiments on SUN RGB-D with the same setup as those on ScanNetV2.
The findings are also similar to those in ScanNetV2.
It can be observed from Tab.~\ref{table2_vsun} that the accuracy of directly transplanted CNNs pruning method is reduced to some extent (mAP@0.25 reduced by 0.41) when FLOPs drop by 58.13\%, while {\cpt} improved the detection model by 0.32\% mAP@0.25 and 1.60\% mAP@0.5 with a FLOPs drop of 61.93\%.
Even when FLOPs drop to 70\%, our method's mAP@0.25 drop only 0.6\%, which is obviously better.

\begin{table}[t]
\centering
\caption{Comparisons of object detection on the ScanNet dataset. The baseline PNN model is GroupFree3D.}\label{table3_gs}
\vspace{-5pt}
\resizebox{\columnwidth}{!}{
\begin{tabular}{lcccc}
\toprule
\textbf{Method} & \textbf{mAP@0.25} & \textbf{mAP@0.50} & \textbf{Params. (K)}      & \textbf{GFLOPs ($\downarrow$\%)}\\ \midrule
Baseline        & 68.22             & 52.61             & 2438.34          & 21.78 (--)                 \\ \cmidrule(lr){1-5}
ResRep          & 68.24             & 51.48             & 1910.73          & 10.90 (49.95)          \\
\textbf{ResRep+{\cpt}}   & \textbf{68.86}    & \textbf{52.08}    & \textbf{1654.35} & \textbf{10.88 (50.05)} \\ \cmidrule(lr){1-5}
ResRep          & 67.21             & 51.28             & 1703.34          & 8.71 (60.01)           \\
\textbf{ResRep+{\cpt}}   & \textbf{68.57}    & \textbf{51.85}    & \textbf{1501.46} & \textbf{8.68 (60.14)}         \\ \bottomrule
\end{tabular}
}
\end{table}

\subsubsection{Evaluation and Comparison on GroupFree3D}

We also conducted experiments on another point-based 3D detection model, GroupFree3D, on ScanNetV2.
Tab.~\ref{table3_gs} summarizes the pruning performance of our approach for GroupFree3D on the ScanNetV2 dataset.
When targeting a moderate compression ratio, our approach can achieve 32.15\% and 50.05\% storage and computation reductions, respectively, with a 0.64\% accuracy increase for mAP@0.25 over the baseline model. In the case of higher compression ratio, {\cpt} still achieves superior performance to other methods.
Specifically, the ResRep loses accuracy by 1.01\% mAP@0.25 when the parameters and flops drop by 30.14\% and 60.01\%, while
in our method, the accuracy increases 0.35\% for mAP@0.25.

\subsection{Ablation Studies}

We conducted ablation studies to validate the Coordinate-Enhancement~(CE) module and the Knowledge-Recycling~(KR) module in {\cpt}.
All results provided in the section are tested on ScanObjectNN with PointNeXt-S~(C=32) as the baseline and HRank as the pruning method.
We evaluated the pruned networks' performance with the FLOPs drop of 75\% and 90\%, and with/without the CE and KR modules to the pruning method.
From Tab.~\ref{ablation_new}, we find that
KR improves OA of 0.84\% and 2.33\% when pruning rate is 75\% and 90\%, respectively, while CE improves OA of 0.32\% and 1.77\% at 75\% and 90\% pruning rates, respectively.
Bringing the two modules together, the OA improvement is 1.84\% and 3.50\% at 75\% and 90\% pruning rates, respectively.
The ablation study results have validated the effectiveness of all designs in {\cpt}.

\begin{table}[t]
\centering
\caption{Ablation study of different components in CP$^3$. Results are of classification on the ScanObjectNN dataset with PointNeXt-S (C=32) as the baseline.
`CE' represents the coordinate-enhanced module, and `KR' represents the knowledge recycling module.}
\vspace{-5pt}
\label{ablation_new}
\resizebox{.85\columnwidth}{!}{
\begin{tabular}{c|c|c|ccc}
\toprule
\textbf{Setting} & \textbf{CE} & \textbf{KR} & \textbf{Pruning Rate} & \textbf{OA}    & \textbf{mAcc}   \\
\midrule
    Baseline & &     & --         & 88.20  & 86.40   \\
\cmidrule(lr){1-6}
HRank &
    &     & 0.75      & 84.79 & 81.93  \\
&    & \checkmark   & 0.75      & 85.63 & 82.97  \\
& \checkmark   &     & 0.75      & 85.11 & 82.13  \\
HRank+{\cpt} &
\checkmark   & \checkmark   & 0.75      & \textbf{86.63} & \textbf{83.63}  \\
\cmidrule(lr){1-6}
HRank &
    &     & 0.90       & 81.33 & 78.32  \\
&    & \checkmark   & 0.90       & 83.66 & 81.32  \\
& \checkmark   &     & 0.90       & 83.10  & 80.47  \\
HRank+{\cpt} &
\checkmark   & \checkmark   & 0.90       & \textbf{84.83} & \textbf{82.74} \\ \bottomrule
\end{tabular}
}
\end{table}

\section{Conclusion}
In this paper, we focus on 3D point-based network pruning and design a 3D channel pruning plug-in (\cpt) that can be used with existing 2D CNN pruning methods.
To the best of our knowledge, this is the first pruning work explicitly considering the characteristics of point cloud data and point-based networks.
Empirically, we show that the proposed {\cpt} is universally effective for a wide range of point-based networks and 3D tasks.

\section*{Acknowledgement}
This work was partially supported by National Natural Science Foundation of China (NSFC) under Grant 62271203,
the Science and Technology Innovation Action Plan of Shanghai under Grant 2251110540,
and Shanghai Pujiang Program under Grant 21PJ1420300.

{\small
\bibliographystyle{ieee_fullname}
\bibliography{ref}
}

\clearpage

\appendix
\twocolumn[{
\begin{center}
\textbf{\Large Supplementary Materials for \\
CP$^3$: Channel Pruning Plug-in for Point Cloud Network}
\end{center}
}]
In the supplementary material, we provide more experimental comparisons on object detection in Sec.~\ref{supp:od} and segmentation tasks in Sec.~\ref{supp:ss} , and we showcase the effectiveness of the proposed knowledge recycling module in Sec.~\ref{supp:kr}, we also analyzed the pruning rates of different layers in Sec.~\ref{supp:al}.

\section{More Experimental Results on 3D Object Detection}
\label{supp:od}
To further illustrate the effectiveness of our proposed \cpt, we incorporated CP$^3$ with pruning methods HRank~\cite{jia2021arank} and CHIP~\cite{sui2021chip} to prune VoteNet~\cite{qi2019deep} on ScanNetV2~\cite{dai2017scannet} and SUN RGB-D~\cite{song2015sun} for 3D object detection.

\begin{table}[h!]
\centering
\caption{Comparisons of object detection performance on the ScanNetV2 dataset.
The baseline PNN model is VoteNet.}\label{tabl1_vsn_supp}
\resizebox{\columnwidth}{!}{
\begin{tabular}{lcccc}
\toprule
\textbf{Method} & \textbf{mAP@0.25} & \textbf{mAP@0.50} & \textbf{Params. (K)} & \textbf{GFLOPs ($\downarrow$\%)}      \\ \midrule
Baseline     & 62.34    & 40.82   & 641.92       & 5.78 (--)     \\ \cmidrule(lr){1-5}
HRank        & 61.04    & 37.99   & 249.82       & 2.46 (57.4)    \\
HRank+CP$^3$ & 61.66    & 39.25   & 239.43       & 2.44 (57.8)    \\
HRank        & 59.46    & 35.98   & 178.16       & 1.87 (67.7)    \\
HRank+CP$^3$ & 60.51    & 39.15   & 169.87       & 1.80 (68.9)    \\ \cmidrule(lr){1-5}
CHIP         & 62.17    & 41.37   & 247.72       & 2.49 (56.9)    \\
CHIP+CP$^3$  & 62.33    & 41.49   & 245.45       & 2.45 (57.6)    \\
CHIP         & 60.86    & 39.94   & 176.88       & 1.89 (67.3)    \\
CHIP+CP$^3$  & 61.55    & 40.43   & 172.78       & 1.87 (67.6)    \\
\bottomrule
\end{tabular}
}
\end{table} 

\para{ScanNetv2}
Tab.~\ref{tabl1_vsn_supp} shows the comparison results of directly applying advanced pruning methods (HRank, CHIP) and implementation them with \cpt.
Overall, CP$^3$ consistently improved the performance of existing advanced CNN pruning methods under different pruning rates.
For instance, in the case of applying HRank with 67.7\% FLOPs reduction, by incorporating \cpt, the mAP@0.50 increased 3.17\% (35.95\% vs. 39.15\%) while achieving 1.2\% more FLOPs reduction (67.7\% vs. 68.9\%).

\begin{table}[h!]
\centering
\caption{Comparisons of object detection performance on the SUN RGB-D dataset. The baseline PNN model is VoteNet.}\label{table2_vsun_supp}
\resizebox{\columnwidth}{!}{
\begin{tabular}{lcccc}
\toprule
\textbf{Method} & \textbf{mAP@0.25} & \textbf{mAP@0.50} & \textbf{Params. (K)} & \textbf{GFLOPs ($\downarrow$\%)}      \\ \midrule
Baseline        & 59.78             & 35.77             & 641.92               & 5.78 (--)            \\  \cmidrule(lr){1-5}
HRank           & 59.22             & 34.26             & 249.82               & 2.46 (57.4)          \\
HRank+CP$^3$    & 60.21             & 34.96             & 245.32               & 2.44 (57.8)           \\
HRank           & 57.68             & 31.30             & 178.88               & 1.87 (67.7)          \\
HRank+CP$^3$    & 59.22             & 33.18             & 176.03               & 1.85 (68.0)           \\ \cmidrule(lr){1-5}
CHIP            & 59.54             & 35.74             & 248.31               & 2.49 (56.9)          \\
CHIP+CP$^3$     & 59.88             & 35.84             & 242.12               & 2.43 (58.0)            \\
CHIP            & 58.63             & 35.07             & 176.23               & 1.89 (67.3)          \\
CHIP+CP$^3$     & 59.13             & 35.32             & 172.02               & 1.87 (67.6)            \\
\bottomrule
\end{tabular}
}
\end{table} 

\para{SUN RGB-D}
We reported the comparison results on the {SUN RGB-D} dataset in Tab.~\ref{table2_vsun_supp}.
For both HRank and CHIP, the implementation with {\cpt} achieved higher accuracy performance with higher FLOPs reduction, similar to our observations on other tasks and datasets.

\begin{figure*}[t]
    \centering
    \includegraphics[width=\linewidth]{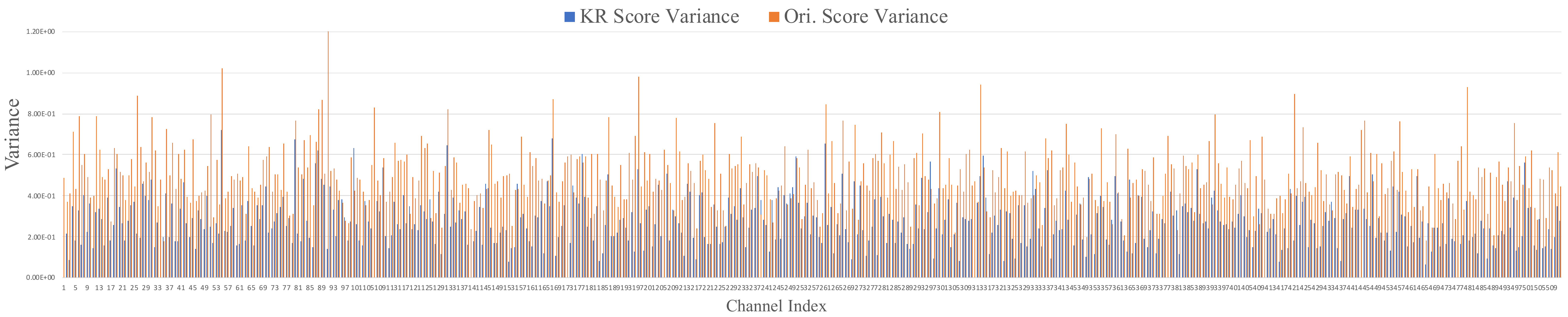}
    \caption{The variance of the importance score of each channel in the 10-th layer of PointNeXt-S. The x-axis represents channel indices, and the y-axis represents the variances of each channel importance scores, which are calculated by 2,468 input samples.}
    \label{variance}
\end{figure*}

\section{More Experimental Results on Semantic Segmentation}
\label{supp:ss}

To investigate the generality of our work, we extended the comparisons on semantic segmentation. We conducted the experiment on the S3DIS~\cite{DBLP:conf/cvpr/ArmeniSZJBFS16} dataset of PointNeXt-S and PointNeXt-XL, and two advanced pruning methods are evaluated.

\begin{table}[h!]
\centering
\caption{Comparisons of semantic segmentation performance on the S3DIS dataset (evaluated in Area-5) with PointNeXt-S~\cite{qian2022pointnext}.}
\label{seg_s3dis_supp_ps}
\resizebox{.95\linewidth}{!}{
\begin{tabular}{l|ccccc}
\toprule
\multirow{2.5}{*}{\textbf{Method}}
           & \multicolumn{5}{c}{\textbf{PointNeXt-S}}  \\ \cmidrule(lr){2-6}
           & \textbf{OA}    & \textbf{mAcc}  & \textbf{mIoU}  & \textbf{Params. (M)} & \textbf{GFLOPs ($\downarrow\%$)}  \\ \midrule
Baseline     &88.20   &70.70   &64.20   &0.80       &3.60 (--)   \\  \cmidrule(lr){1-6}
HRank        &85.89  &67.27   &60.49  &0.33     &1.53 (57.5)  \\
HRank+{\cpt} &86.18  &67.65   &61.04  &0.31     &1.52 (57.8)  \\
HRank        &84.92  &65.37   &58.73  &0.17     &0.77 (78.6)  \\
HRank+{\cpt} &85.12  &67.48   &60.12  &0.16     &0.74 (79.4)  \\ \cmidrule(lr){1-6}
CHIP         &84.45  &67.72   &61.16  &0.32     &1.53 (57.5)  \\
CHIP+{\cpt}  &84.53  &70.52   &63.62  &0.33     &1.48 (58.9)  \\
CHIP         &84.39  &67.29   &60.63  &0.15     &0.74 (79.4)  \\
CHIP+{\cpt}  &85.04  &69.02   &61.45  &0.16     &0.73 (79.7)  \\ \bottomrule
\end{tabular}}
\end{table}

\para{PointNext-S}
Compared to other PointNeXt zoos, besides the fewer parameters, PointNeXt-S is designed with no InvResMLP blocks and is a simpler network architecture.
The comparison result in Tab.~\ref{seg_s3dis_supp_ps} showed that the consistent outperformance of {\cpt} compared to directly using HRank and CHIP \textbf{without} \cpt.
In the case of applying CHIP with 57.5\% FLOPs reduction, by incorporating \cpt, the mAcc increases 2.8 \% while achieving 1.4 \% more FLOPs reduction.
The result indicated that even for a simpler network architecture, the accuracy performance degradation still occurred when directly implementation of CNN pruning methods, verifying the necessity of the implementation with \cpt.

\begin{table}[h!]
\centering
\caption{Comparisons of semantic segmentation performance on the S3DIS dataset (evaluated in Area-5) with PointNeXt-XL~\cite{qian2022pointnext}.}
\label{seg_s3dis_supp}
\resizebox{.95\linewidth}{!}{
\begin{tabular}{l|ccccc}
\toprule
\multirow{2.5}{*}{\textbf{Method}}
           & \multicolumn{5}{c}{\textbf{PointNeXt-XL}}  \\ \cmidrule(lr){2-6}
           & \textbf{OA}    & \textbf{mAcc}  & \textbf{mIoU}  & \textbf{Params. (M)} & \textbf{GFLOPs ($\downarrow\%$)}  \\ \midrule
Baseline     &91.00   &77.20    &71.10   &41.60   &84.80 (--)         \\  \cmidrule(lr){1-6}
HRank        &90.02   &74.50   &68.22  &13.03  &26.71 (68.5)      \\
HRank+{\cpt} &90.80   &75.85   &69.98  &12.57  &25.78 (69.6)      \\
HRank        &89.79   &74.40   &68.09  &8.80   &18.05 (78.7)      \\
HRank+{\cpt} &90.48   &74.45   &68.41  &8.42   &17.37 (79.5)      \\ \cmidrule(lr){1-6}
CHIP         &89.98   &75.43   &69.21  &17.54  &35.97 (57.6)      \\
CHIP+{\cpt}  &90.57   &75.90   &69.40  &17.00  &34.68 (59.1)      \\
CHIP         &89.15   &74.54   &68.07  &8.42   &17.37 (79.5)      \\
CHIP+{\cpt}  &90.03   &74.61   &68.26  &8.05   &16.62 (80.4)      \\ \bottomrule
\end{tabular}}
\end{table}

\para{PointNext-XL}
We further investigated on the more complex and larger network PointNext-XL. Similar observations on the improvement by {{\cpt}} can be found in Tab.~\ref{seg_s3dis_supp}.
For instance, our approach can achieve 69.8 \% and 69.9  \% storage and computation reductions, respectively, with a 1.3 \% and 1.7\% accuracy increase for mAcc and mIoU over the baseline model.

\begin{table}[h]
\centering
\caption{Comparisons on the SemanticKITTI with RandLA-Net.}
\label{tab_2}
\resizebox{0.42\textwidth}{!}{
\begin{tabular}{l|ccc}
\toprule
\multicolumn{1}{l|}{\multirow{1}{*}{Method}}
& \multicolumn{1}{c}{mIoU} & \multicolumn{1}{c}{Params. (M)} & \multicolumn{1}{c}{FLOPs (\%)}
 \\ \midrule
Baseline (RandLA-Net)          &50.30   &0.95     &100                   \\ \cmidrule{1-4}
CHIP               &49.12  &0.20    &21.7                   \\
CHIP+CP$^3$        &50.21  &0.18    &19.8 \\ \bottomrule
\end{tabular}}
\end{table}

\para{Outdoor Experiments}
{\cpt} focuses on point-based networks (PNNs), and by following prevailing PNN works, we have experimented on the popular large-scale datasets such as ScanObjectNN and S3DIS and achieved promising results in the paper. To further illustrate the validity of our method, we experimented on a outdoor dataset (SemanticKITTI) with RandLA-Net and CHIP. The results are shown in Tab.~\ref{tab_2}.

\section{Exploration on Knowledge Recycling}
\label{supp:kr}

In this section, we took a deeper look into the Knowledge Recycling~(KR) module.
To verify the effectiveness of KR, we performed pruning methods and statistically analyzed the positive effect of KR.
We took the PointNeXt-S with dataset ModelNet40 as an example.
We performed the comparison between directly implementing HRank and HRank with \cpt.
And we focus on the KR scores generated by the KR module (with discarded points) and HRank scores \textbf{without} KR module.
We calculated the variances of scores to justify the robustness of \cpt.
Fig.~\ref{variance} shows the statistics comparison results on the 10-th layer with 512 channels on 2468 test meshed CAD models.
Among 512 channels in the 10-th layer, the percentage is 93.2\% in the case of the KR score variances lower than the original score variances. For instance, in the case of the 25-th channel, the KR score variance is much lower than the original score variance (0.21 vs. 0.89).
Similar results can be found on other layers. These results verify that the KR module enabled the channel importance calculation to be more stable and robust.

\section{Analysis of the layer-wise redundancy}
\label{supp:al}
\begin{figure}[t]
\centering
\includegraphics[width=0.45\textwidth]{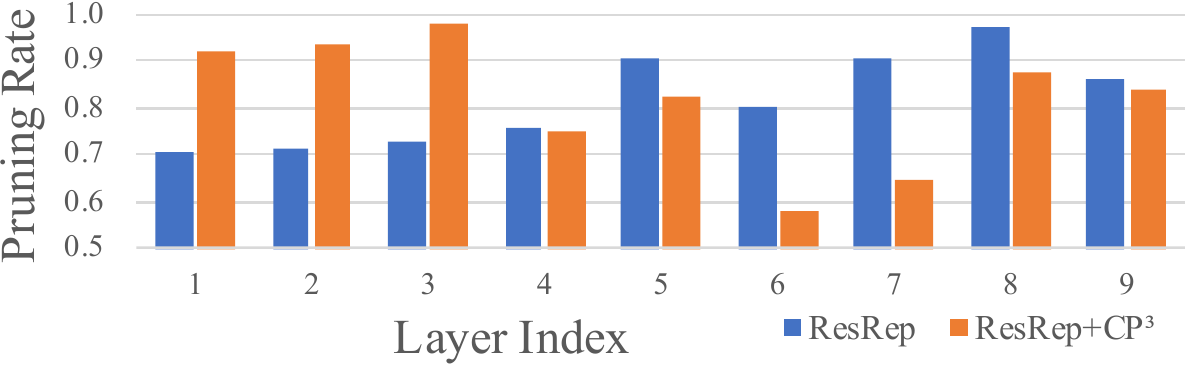}
\caption{Different layer pruning rate. } \label{fig_channel}
\end{figure}

We take the pruned PointNet++ on ScanObjectNN as an example and show the pruning rates for each layer in Fig.~\ref{fig_channel}.
Pruning with {\cpt} eliminates more redundancy on shallow layers and less on $6$th and $7$th layers.
{\cpt} on ResRep achieved higher accuracy (84.80\% vs 83.79\%) with higher FLOPs reduction (84.8\% vs 83.0\%), indicating pruning with CP$^3$ effectively identifies the redundancy in point-based networks.

\end{document}